\documentclass[11pt,letterpaper]{article}
\usepackage{naaclhlt2013}
\usepackage{times}
\usepackage{latexsym}
\usepackage{comment,graphicx,multirow,url}
\title{Authorship Attribution Using Word Network Features}

\author{Shibamouli Lahiri\qquad Rada Mihalcea\\
	    Computer Science and Engineering\\
	    University of North Texas\\
	    Denton, TX 76207, USA\\
	    {\tt shibamoulilahiri@my.unt.edu, rada@cs.unt.edu}}

\date{}

\begin{document}
\maketitle
\begin{abstract}
In this paper, we explore a set of novel features for authorship attribution of documents. These features are derived from a word network representation of natural language text. As has been noted in previous studies, natural language tends to show complex network structure at word level, with low degrees of separation and scale-free (power law) degree distribution. There has also been work on authorship attribution that incorporates ideas from complex networks. The goal of our paper is to explore properties of these complex networks that are suitable as features for machine-learning-based authorship attribution of documents. We performed experiments on three different datasets, and obtained promising results.
\end{abstract}

\section{Introduction}
\label{sec:intro}

Complex networks are ubiquitous~\cite{citeulike:44_cn,Newman01_modularity,Newman:2010:NI:1809753}. Gene regulatory networks~\cite{citeulike:1087985,Vaidehi2001}, social networks~\cite{citeulike:4041910,easley2010} and the Internet~\cite{Park:2005:ILC:993523} are three well-known examples. These networks have certain common properties, viz. a small-world structure, core-periphery property, shrinking diameter, scale-free (power law) degree distribution and low degrees of separation~\cite{Leskovec:2010:KGA:1756006.1756039}.

Complex network structure has recently been observed in natural language as well~\cite{canchosole_2001,Liang20094901,MDLD02}. Networks of words constructed from text documents show small-world structure, scale-free (power law) degree distribution and a low degree of separation~\cite{Matsuo:2001:DSW:645654.665684,Matsuo:2001:KEK:647858.738697}. Such word networks have seen applications in text genre identification~\cite{DBLP:journals/corr/abs-1007-3254}, web query analysis~\cite{roy2011complex}, language models~\cite{coling2012_semantics_cn}, and opinion mining~\cite{opinion_cn_recent}.

In this paper, we explore a novel set of features for the traditional NLP task of authorship attribution. These features are derived from a word network representation of documents (cf. Section~\ref{sec:cn}). Note that the complex networks approach for authorship attribution is not new by itself (Section~\ref{sec:related}), but the existing body of work has not yet thoroughly explored the usefulness of machine learning on complex network features for authorship attribution. And this is the gap we want to bridge in our work. We extracted network features from different documents, trained classifiers on documents of known authorship, and tested them on documents of unknown authorship.

We have focused on three different datasets in this paper (Section~\ref{sec:data}), two of which are from authorship attribution competitions. An array of experiments on these datasets (Section~\ref{sec:exp}) justifies the speculation that complex network features are indeed useful for classification-based authorship attribution, albeit we need to be careful in choosing different types of features. We discuss in some detail (Section~\ref{sec:exp}) the good and not-so-good word network features for authorship attribution, along with possible reasons. A set of future directions has been outlined in Section~\ref{sec:conclusion}.

\section{Related Work}
\label{sec:related}

% Describe work in authorship attribution and AA with CN

While authorship attribution is a well-known problem in NLP (see, e.g., the surveys by Juola~\cite{Juola:2006:AA:1373450.1373451}, Stamatatos~\cite{Stamatatos:2009:SMA:1527090.1527102}, and Koppel et al~\cite{Koppel:2009:CMA:1484611.1484627}), complex networks have only recently been applied to the authorship attribution problem.

An important take in this line of research is due to~\cite{DBLP:conf/semco/ArunSM09}, where the authors constructed weighted networks of stopwords found in a document. Authorship of test documents was determined by ranking test documents based on their graph similarity with stopword networks of training documents.

\cite{Antiqueira2006a} were among the first to investigate the possibility of authorship attribution using complex network features. They used clustering coefficient, \emph{component dynamics deviation} and \emph{degree correlation} of word networks to successfully classify documents according to their authorship. On the other hand,~\cite{6007535_fiqi} worked on the problem of identifying \emph{translator style} from word network properties. They used small \emph{motifs} of size three and their distributions as signatures of translators. Note that motifs are local subgraphs carrying important neighborhood information which is often lost in global graph properties like diameter and reciprocity. We will come back to this issue again in Section~\ref{sec:exp}.

\cite{Mehri20122429} defined a probability measure called \emph{attribution probability} for authorship identification in Persian language. They obtained good accuracy in authorship classification of Persian books using the power-law exponent of word networks of those books and the so-called \emph{non-extensivity measure} (q-parameter) of the networks.

Finally,~\cite{amancio2011comparing} discovered links between intermittency (burstiness) in word distribution and topological properties of word networks. They found that skewness in word intermittency distribution and average length of shortest paths have greater dependency on the authorship of a particular document. They also used machine learning to classify documents according to authorship. The best results were obtained using nearest neighbor algorithm (65\% accuracy).

While all the above studies are extremely important, we have found none that systematically looked into different types of word network features and tried to identify what works best and when. Moreover, most of the current work in authorship attribution using word network features have been done by researchers in Physics, and the idea of complex networks has only recently begun to see acceptance in Computer Science, especially NLP~\cite{coling2012_semantics_cn}. While the marriage between two seemingly disparate domains like authorship attribution and complex networks may sound unintuitive at first blush, note that networks have long been a part of mainstream NLP and Information Retrieval~\cite{mihalcea2011graph}. It is only very recently that word networks have started seeing application in stylometry. Current approaches, however, are plagued by small training and test datasets, lack of evaluation on publicly available authorship attribution competition data, and a lack of clarity regarding where word network features stand in comparison with standard authorship attribution features like stopword term frequency. We aim to bridge this gap in this paper.

\section{Complex Networks of Text}
\label{sec:cn}

\subsection{Word Networks}
\label{subsec:wn}

We have so far used the term ``word networks'' rather loosely. Let us now make its definition more concrete. In a nutshell, a ``word network'' of a particular document is a network (graph) of unique words found in that document. Each node (vertex) in this network is a word. Edges between two nodes (unique words) can be constructed in several different ways. The simplest type of edge connects word A to word B, if word A is immediately followed by word B in the document at least once. In our work, we have assumed a directed edge with direction from word A to word B. Note that we could have used undirected edges as well (cf.~\cite{mihalcea-tarau:2004:EMNLP}). Moreover, edges can be weighted/unweighted. We assumed unweighted edges.

A deeper issue with the construction of word networks concerns what we should do with stopwords. Should we keep them, or should we remove them? Since stopwords and function words have proved to be of special importance in previous authorship attribution studies~\cite{Diederich:2003:AAS:776973.776982,menon-choi:2011:RANLP}, we chose to keep them in our word networks. But unlike~\cite{DBLP:conf/semco/ArunSM09}, who used only stopwords to construct their word networks, we used all the unique words in a document.

Two other choices we made in the construction of our word networks concern sentence boundaries and word co-occurrence. Word networks can be constructed either by respecting sentence boundaries (where the last word of sentence 1 does \emph{not} link to the first word of sentence 2), or by disregarding them. In our case, we disregarded all sentence boundaries. Moreover, a network edge can either link two words that appeared side-by-side in the original document, or it can link two words that appeared within a window of $n$ words in the document. In our case, we chose the first option - linking unique words that appeared side-by-side at least once. Finally, we did not perform any stemming/morphological analysis to retain subtle cues that might be revealed from inflected/derived words.

\begin{figure}
\centering
\includegraphics[width=0.9\linewidth]{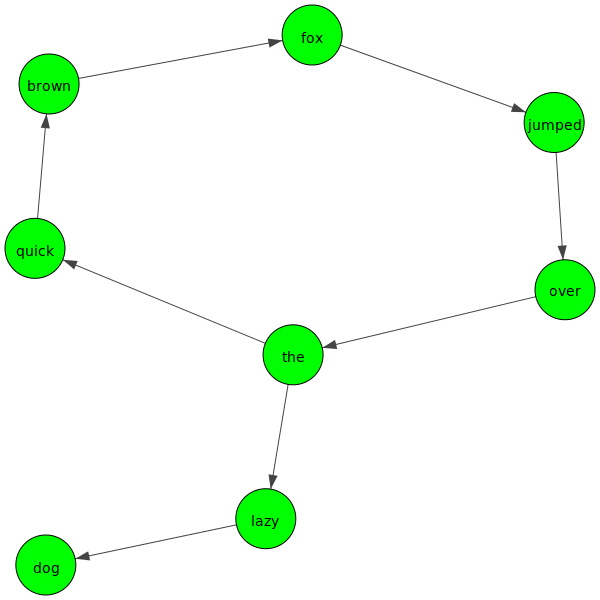}
\caption{Word network of the sentence \emph{``the quick brown fox jumped over the lazy dog''}.}
\label{fig:cn_quick}
\end{figure}

With the above-mentioned choices, the word network of an example sentence (``the quick brown fox jumped over the lazy dog'') is shown in Figure~\ref{fig:cn_quick}. Note that the word ``the'' appeared twice in this sentence, so the corresponding network contains a cycle that starts at ``the'' and ends at ``the''. In a realistic word network of a large document, there can be many such cycles. In addition, it is observed that such word networks show power-law degree distribution and a small-world structure~\cite{canchosole_2001,Matsuo:2001:DSW:645654.665684}.

There can be several variations on word networks. In a more basic variant, words are replaced with characters and character n-grams~\cite{kevselj2003n,Giannakopoulos:2008:SSE:1410358.1410359,escalante2011local}. In principle, we could also use phrases, clauses, sentences or even paragraphs instead of words~\cite{Ohsawa:1998:KAI:582987.785950,mihalcea-tarau:2004:EMNLP}. Note, however, that as we increase the granularity of individual nodes, the resulting network starts losing subtleties (e.g., cycles) that are apparent at word level.

% Extracted features - global (i.e., summary) features vs local features (word lists + top freq words)

\subsection{Network Features}
\label{subsec:wn_features}

Once the word networks have been constructed, we extract a set of simple features from these networks.\footnote{We used the igraph~\cite{igraph_cite} software package for graph feature extraction.} Some of these features represent global properties of a particular network; we call them ``summary features''. Others represent local properties of individual nodes; we call them ``local features''. We have extracted 127 summary features for each graph, and ten local features for each node in a graph. The summary features are:

\begin{enumerate}
\item number of vertices ($|$V$|$)
\item number of edges ($|$E$|$)
\item is the graph strongly connected? (True/False)
\item number of strongly connected components
\item number of articulation points (\emph{cut vertices})
\item global clustering coefficient of the graph
\item adhesion, cohesion, assortativity and density (with and without self-loops) of the graph
\item densification power law exponent or \emph{shrinking exponent} (log $|$E$|$/log$|$V$|$)~\cite{Leskovec:2010:KGA:1756006.1756039}
\item four types of reciprocity measures, with and without self-loops
\item girth (length of the shortest cycle in the graph, ignoring self-loops)
\item statistical properties of in-degree, out-degree and degree distribution (minimum, maximum, average, median, 25th and 75th percentiles, variance, standard deviation, skewness, kurtosis, fitted $\alpha$ (\emph{power-law exponent}))
\item statistical properties of local clustering coefficient distribution (as described above)
\item statistical properties of in-coreness, out-coreness and coreness distribution (as described above)
\item statistical properties of in-neighborhood size (order 1), out-neighborhood size (order 1) and neighborhood size (order 1) distribution (as described above)
\end{enumerate}

Each document is represented as a \emph{summary feature vector} of length 127. The local features, on the other hand, are defined over \emph{nodes} (unique words) in the graph. They are:

\begin{enumerate}
\item in-degree, out-degree and degree
\item in-coreness, out-coreness and coreness
\item in-neighborhood size (order 1), out-neighborhood size (order 1) and neighborhood size (order 1)
\item local clustering coefficient
\item term frequency of the corresponding word in the document (\emph{baseline feature set}\footnote{Please see Section~\ref{sec:exp} for details.})
\end{enumerate}

We take a set of \emph{representative words}, and convert a document into a local feature vector - each local feature pertaining to one word in the set of representative words. For example, when we use top 200 most frequent words as the representative set, a document can be represented as the degree vector of these 200 words in the document's word network, or as the local clustering coefficient vector of these words in the word network, or as the coreness vector of the words (and so on). A document can also be represented as a concatenation (\emph{mixture}) of these vectors. For example, it can be represented as $concat(degree\_vector, coreness\_vector)$ of top 200 most frequent words. We are yet to explore how such mixed feature sets perform in authorship classification. It constitutes a part of our future work (Section~\ref{sec:conclusion}).

We experimented with the following lists of words as our representative word-set:

\begin{enumerate}
\item two lists of stopwords from~\cite{menon-choi:2011:RANLP} (174 and 667 stopwords, respectively)
\item a list of 123 mood words from~\cite{menon-choi:2011:RANLP}
\item a publicly available list of 571 common words\footnote{\url{http://www.cse.unt.edu/~rada/CSCE5200/Resources/common_words}}
\item a publicly available list of 54 stopwords\footnote{\url{http://www.cse.unt.edu/~rada/CSCE5200/Resources/stopwords}}
\item top k most frequent words with k = 100, 200, 500, 1000, on the whole corpus (cf. Section~\ref{sec:data})
\end{enumerate}

\section{Data}
\label{sec:data}

We have collected three datasets for our experiments. The first one is a collection of 3,036 English e-books from Project Gutenberg digital library\footnote{\url{http://www.gutenberg.org/}}. These books were written by 142 authors. We specifically chose not to use translated books (i.e., books translated from other languages to English), because previous research indicates that translator's style leaves distinct footprints on the style of the original author~\cite{6007535_fiqi,veni_translat}, thereby corrupting the original style of a book.

This dataset has class imbalance. Some authors have a large number of books (e.g., Anthony Trollope has 71 books), some have a medium number of books (e.g., Abraham Lincoln has 16 books), and yet some others have a small number of books (e.g., Ezra Pound has 2 books). The distribution is right-skewed with skewness 1.4 and kurtosis 1.34. On average, there are 21.4 books per author, with a standard deviation of 22.35 books. 

After downloading the e-books from Project Gutenberg website, we manually cleaned them to remove metadata, license information and transcribers' notes. Then we created word networks from these books, extracted network features as described in the previous section, and created a vocabulary of most frequent words found in the whole corpus. These most frequent words were used later in our authorship classification experiments (cf. Section~\ref{sec:exp}).

Apart from Project Gutenberg, we used two other publicly available datasets from past authorship attribution competitions - the Ad-hoc Authorship Attribution Competition (AAAC) dataset~\cite{Juola:2006:AA:1373450.1373451}, and the PAN 2012 (PAN12) traditional authorship attribution task dataset\footnote{\url{http://www.uni-weimar.de/medien/webis/research/events/pan-12/pan12-web/authorship.html}}. These datasets have training and test materials for open-class and closed-class authorship attribution problems. We used the closed-class data, as the open-class problems are out of scope of our present study. There are ten closed-class problems for the AAAC dataset (problems A, B, C, G, H, I, J, K, L, M), and three closed-class problems for the PAN12 dataset (problems A, C, I). For each closed-class problem, we have a set of training documents with known authorship, and a set of test documents with unknown authorship. For evaluation purposes, ground truth authorship class labels for the test data were provided by contest-organizers. These datasets have the number of training documents balanced between different authors (except AAAC problem K).

\section{Experiments}
\label{sec:exp}

% Explain the need for stratified cross-validation in weka (class imbalance in gutenberg dataset)

Recall from Section~\ref{subsec:wn_features} that we extracted two types of features from our word networks - the \emph{summary features} and the \emph{local features}. With these features, we performed a series of classification experiments.\footnote{We used Weka~\cite{Hall:2009:WDM:1656274.1656278} for all our classification experiments.}

\subsection{On Project Gutenberg Data}
\label{subsec:gutenberg}

We cast the problem of authorship attribution as a multiclass classification of 3,036 documents among 142 authors. Since there is class imbalance in the Gutenberg data we collected (cf. Section~\ref{sec:data}), in this section we report results of stratified 10-fold cross-validation with Weka.

\begin{table}
\begin{center}
\footnotesize
\begin{tabular}{lc}
\hline
\textbf{Classifier} & \textbf{Cross-validation Accuracy (\%)}\\
\hline
1NN & 30.86\\
%2NN & 27.54\\
%3NN & 28.72\\
J48 & 28.42\\
Na\"{i}ve Bayes & 17.32\\
OneR & 13.87\\
SVM & 28.82\\
AdaBoost & 5.17\\
\textbf{Logit Boost} & \textbf{34.82}\\
Random Baseline & 0.7 ($\pm$ 0.15)\\
Majority Class Baseline & 3.2\\
\hline
\end{tabular}
\end{center}
\caption{\label{tab:gutenberg_summary_multiclass}Multiclass classification of Project Gutenberg documents using summary features. For random basline, we show the mean accuracy over 100 runs and its standard deviation (in parenthesis).}
\end{table}

\begin{table}
\begin{center}
\footnotesize
\begin{tabular}{lc}
\hline
\textbf{Classifier} & \textbf{Cross-validation Accuracy (\%)}\\
\hline
1NN & 30.86\\
J48 & 26.45\\
%PART & 28.13\\
Na\"{i}ve Bayes & 4.81\\
OneR & 5.57\\
SVM & 3.69\\
AdaBoost & 26.52\\
\textbf{Logit Boost} & \textbf{33.53}\\
Random Baseline & 0.7 ($\pm$ 0.15)\\
Majority Class Baseline & 0.032\\
\hline
\end{tabular}
\end{center}
\caption{\label{tab:gutenberg_summary_onevsall}One-vs-all classification of Project Gutenberg documents using summary features. For random basline, we show the mean accuracy over 100 runs and its standard deviation (in parenthesis).}
\end{table}

\begin{table}
\begin{center}
\footnotesize
\begin{tabular}{lc}
\hline
\textbf{Top $k$} & \textbf{Best Cross-validation Accuracy (\%)}\\
\hline
$k$ = 100 & 66.10 (1NN)\\
$k$ = 200 & 71.34 (1NN)\\
$k$ = 500 & 75.33 (1NN)\\
\textbf{$k$ = 1000} & \textbf{76.78 (1NN)}\\
\hline
\end{tabular}
\end{center}
\caption{\label{tab:gutenberg_most_frequent_multiclass}Multiclass classification of Project Gutenberg documents using term frequency of top-$k$ most frequent words. Stratified ten-fold cross-validation results are shown, along with the classifiers that achieved these results.}
\end{table}

Table~\ref{tab:gutenberg_summary_multiclass} and Table~\ref{tab:gutenberg_summary_onevsall} show the results of using graph summary features on different classifiers in Weka. Table~\ref{tab:gutenberg_summary_onevsall} uses a different flavor of multiclass classification known as \emph{one-vs-all classification}~\cite{Rifkin:2004:DOC:1005332.1005336}. Note that in both cases, graph summary features perform substantially better than a na\"{i}ve random baseline where documents are randomly assigned to authors, and a majority class baseline, where all documents are assigned to the author of the highest number of books. Logit Boost performs the best among all classifiers in both cases.

Although the performance of graph summary features is better than random baseline and majority class baseline, they fail miserably when we compare them against stronger baselines like term frequency of most frequent words (Table~\ref{tab:gutenberg_most_frequent_multiclass}). In Table~\ref{tab:gutenberg_most_frequent_multiclass}, we show the results of using top 100, 200, 500 and 1000 most frequent words' term frequency as features\footnote{Note that these most frequent words were extracted from the whole corpus (Section~\ref{sec:data}).}. We experimented with several different classifiers\footnote{1NN, 2NN, 3NN, J48, Na\"{i}ve Bayes, OneR and AdaBoost.}, but show only the best results in Table~\ref{tab:gutenberg_most_frequent_multiclass} for the sake of brevity. All best results were obtained with 1NN (1-nearest neighbor) classifier.

It was observed that if we combine graph summary features with term frequency of most frequent words, the result (best CV accuracy of 41.70\% using 1NN) tends to get much worse than the one obtained with term frequency only. This observation led us to believe that graph summary features may not be very useful for authorship attribution, whatever classifier we may choose to use. On the other hand, this also indicates that \emph{local features} may hold more information and subtle cues than summary features, which may help in authorship classification\footnote{Recall that \emph{local features} are features extracted on representative sets of words (cf. Section~\ref{subsec:wn_features}).}.

\begin{table*}
\begin{center}
\footnotesize
\begin{tabular}{cccccc}
\hline
\multirow{4}{*}{\textbf{Local Feature}} & \multicolumn{5}{c}{\textbf{Best Cross-validation Accuracy (\%) on Representative Sets of Words}}\\
& & & & &\\
& \multirow{2}{*}{174 stopwords} & bigger list of & \multirow{2}{*}{123 mood words} & \multirow{2}{*}{571 common words} & second list of\\
& & 667 stopwords & & & 54 stopwords\\
\hline
Term Frequency & \multirow{2}{*}{60.97 (1NN)} & \multirow{2}{*}{70.12 (1NN)} & \multirow{2}{*}{46.51 (Na\"{i}ve Bayes)} & \multirow{2}{*}{71.87 (1NN)} & \multirow{2}{*}{58.63 (1NN)}\\
(baseline) & & & & &\\
Clustering Coefficient & 53.46 (Na\"{i}ve Bayes) & 42.85 (Na\"{i}ve Bayes) & 15.05 (Na\"{i}ve Bayes) & 44.96 (Na\"{i}ve Bayes) & 54.25 (1NN)\\
Degree & 69.86 (1NN) & 78.66 (1NN) & 48.35 (1NN) & 78.66 (1NN) & 62.55 (1NN)\\
Coreness & 57.81 (1NN) & \textbf{78.85 (1NN)} & \textbf{50.26 (1NN)} & \textbf{78.89 (1NN)} & 33.79 (1NN)\\
Neighborhood Size & \multirow{2}{*}{\textbf{69.96 (1NN)}} & \multirow{2}{*}{78.29 (1NN)} & \multirow{2}{*}{48.02 (1NN)} & \multirow{2}{*}{78.62 (1NN)} & \multirow{2}{*}{\textbf{63.21 (1NN)}}\\
(order 1) & & & & &\\
\hline
\end{tabular}
\end{center}
\caption{\label{tab:gutenberg_stop_multiclass}Multiclass classification of Project Gutenberg documents using local features on representative sets of words. Stratified ten-fold cross-validation results are shown, along with the classifiers that achieved these results. Best results in different columns are boldfaced.}
\end{table*}

\begin{table*}
\begin{center}
\footnotesize
\begin{tabular}{cccccc}
\hline
\textbf{Local Feature} & \multicolumn{5}{c}{\textbf{Best Cross-validation Accuracy (\%) on Representative Sets of Words}}\\
+ & & & & &\\
Term Frequency & \multirow{2}{*}{174 stopwords} & bigger list of & \multirow{2}{*}{123 mood words} & \multirow{2}{*}{571 common words} & second list of\\
(baseline) & & 667 stopwords & & & 54 stopwords\\
\hline
Clustering Coefficient & 63.04 (SVM) & 68.02 (Logit Boost) & 41.86 (Logit Boost) & 68.71 (Logit Boost) & 56.16 (1NN)\\
Degree & 68.41 (1NN) & 78.03 (1NN) & 52.73 (SVM) & 78.23 (1NN) & 62.75 (1NN)\\
Coreness & 62.94 (1NN) & \textbf{78.23 (1NN)} & \textbf{54.84 (SVM)} & \textbf{78.75 (1NN)} & 50.30 (1NN)\\
Neighborhood Size & \multirow{2}{*}{\textbf{68.84 (1NN)}} & \multirow{2}{*}{77.73 (1NN)} & \multirow{2}{*}{54.74 (SVM)} & \multirow{2}{*}{77.83 (1NN)} & \multirow{2}{*}{\textbf{63.18 (1NN)}}\\
(order 1) & & & & &\\
\hline
\end{tabular}
\end{center}
\caption{\label{tab:gutenberg_stop_multiclass_mix}Multiclass classification of Project Gutenberg documents using local features and term frequency on representative sets of words. Stratified ten-fold cross-validation results are shown, along with the classifiers that achieved these results. Best results in different columns are boldfaced.}
\end{table*}

\begin{table*}
\begin{center}
\footnotesize
\begin{tabular}{ccccc}
\hline
\multirow{3}{*}{\textbf{Local Feature}} & \multicolumn{4}{c}{\textbf{Best Cross-validation Accuracy (\%) on Top $k$ Most Frequent Words}}\\
& & & &\\
& $k$ = 100 & $k$ = 200 & $k$ = 500 & $k$ = 1000\\
\hline
Term Frequency & \multirow{2}{*}{65.61 (1NN)} & \multirow{2}{*}{70.75 (1NN)} & \multirow{2}{*}{75.03 (1NN)} & \multirow{2}{*}{76.61 (1NN)}\\
(baseline) & & & &\\
Clustering Coefficient & 66.27 (1NN) & 71.57 (SVM) & 67.09 (SVM) & 59.82 (Logit Boost)\\
Degree & 69.66 (1NN) & 74.31 (1NN) & 78.79 (1NN) & 80.50 (1NN)\\
Coreness & 44.93 (1NN) & 64.00 (1NN) & \textbf{80.67 (SVM)} & 79.97 (1NN)\\
Neighborhood Size & \multirow{2}{*}{\textbf{70.09 (1NN)}} & \multirow{2}{*}{\textbf{74.44 (1NN)}} & \multirow{2}{*}{79.25 (1NN)} & \multirow{2}{*}{\textbf{80.80 (1NN)}}\\
(order 1) & & & &\\
\hline
\end{tabular}
\end{center}
\caption{\label{tab:gutenberg_mf_multiclass}Multiclass classification of Project Gutenberg documents using local features on most frequent words. Stratified ten-fold cross-validation results are shown, along with the classifiers that achieved these results. Best results in different columns are boldfaced.}
\end{table*}

\begin{table*}
\begin{center}
\footnotesize
\begin{tabular}{ccccc}
\hline
\textbf{Local Feature} & \multicolumn{4}{c}{\textbf{Best Cross-validation Accuracy (\%) on Top $k$ Most Frequent Words}}\\
+ & & & &\\
Term Frequency (baseline) & $k$ = 100 & $k$ = 200 & $k$ = 500 & $k$ = 1000\\
\hline
Clustering Coefficient & 66.80 (SVM) & 74.34 (SVM) & 66.11 (Logit Boost) & 68.84 (Logit Boost)\\
Degree & 69.43 (1NN) & 74.34 (1NN) & 78.46 (1NN) & 80.30 (1NN)\\
Coreness & 59.16 (1NN) & 69.01 (1NN) & 78.00 (1NN) & 80.11 (1NN)\\
Neighborhood Size & \multirow{2}{*}{\textbf{69.80 (1NN)}} & \multirow{2}{*}{\textbf{74.74 (1NN)}} & \multirow{2}{*}{\textbf{78.85 (1NN)}} & \multirow{2}{*}{\textbf{80.63 (1NN)}}\\
(order 1) & & & &\\
\hline
\end{tabular}
\end{center}
\caption{\label{tab:gutenberg_mf_multiclass_mix}Multiclass classification of Project Gutenberg documents using local features and term frequency on most frequent words. Stratified ten-fold cross-validation results are shown, along with the classifiers that achieved these results. Best results in different columns are boldfaced.}
\end{table*}

\begin{table*}
\begin{center}
\footnotesize
\begin{tabular}{lcc}
\hline
\textbf{Rank} & \textbf{Word Network Feature} & \textbf{Information Gain}\\
\hline
1 & Term frequency of the word \emph{until} & 0.621 \\
2 & Neighborhood size of the word \emph{until} & 0.611 \\
3 & Degree of the word \emph{until} & 0.610 \\
4 & Neighborhood size of the word \emph{by} & 0.576 \\
5 & Term frequency of the word \emph{several} & 0.574 \\
6 & Term frequency of the word \emph{thus} & 0.555 \\
7 & Degree of the word \emph{thus} & 0.553 \\
8 & Degree of the word \emph{several} & 0.544 \\
9 & Neighborhood size of the word \emph{several} & 0.543 \\
10 & Coreness of the word \emph{thus} & 0.538 \\
11 & Neighborhood size of the word \emph{though} & 0.524 \\
12 & Term frequency of the word \emph{had} & 0.509 \\
13 & Term frequency of the word \emph{by} & 0.507 \\
14 & Neighborhood size of the word \emph{may} & 0.505 \\
15 & Degree of the word \emph{or} & 0.499 \\
16 & Clustering coefficient of the word \emph{said} & 0.497 \\
17 & Coreness of the word \emph{upon} & 0.489 \\
18 & Coreness of the word \emph{whom} & 0.489 \\
19 & Degree of the word \emph{by} & 0.488 \\
20 & Neighborhood size of the word \emph{returned} & 0.484 \\
\hline
\end{tabular}
\end{center}
\caption{\label{tab:ig_ranking_gutenberg_500_most_frequent_words}Ranking of term frequency and local word network features based on Information Gain, on Gutenberg data. We took 500 most frequent words on the whole dataset, and collected their term frequency, clustering coefficient, neighborhood size, coreness and vertex degree (for each document) in a single file. This ranking reflects the top 20 among 2,500 features in that file, along with their information gain values. Note that both term frequency as well as local word network features appeared at the top. Moreover, stopwords like \emph{until}, \emph{by}, \emph{several} and \emph{thus} are found to be important predictors of writing style.}
\end{table*}

\begin{table*}
\begin{center}
\footnotesize
\begin{tabular}{lcccc}
\hline
\multirow{3}{*}{\textbf{Dataset}} & \textbf{Best Test Set} & \textbf{Best Test Set} & \multirow{2}{*}{\textbf{Best Reported}} & \multirow{2}{*}{\textbf{Lowest Reported}}\\
& \textbf{Accuracy (\%)} & \textbf{Accuracy (\%)} & \multirow{2}{*}{\textbf{Accuracy (\%)}} & \multirow{2}{*}{\textbf{Non-zero Accuracy (\%)}}\\
& \textbf{w/o term frequency} & \textbf{w/ term frquency} & &\\
\hline
AAAC Problem A & 61.54 & 69.23 & \textbf{84.62} & 15.38\\
AAAC Problem B & \textbf{61.54} & \textbf{61.54} & 53.85 & 7.69\\
AAAC Problem C & \textbf{100.00} & \textbf{100.00} & \textbf{100.00} & 33.33\\
AAAC Problem G & \textbf{100.00} & \textbf{100.00} & 75.00 & 25.00\\
AAAC Problem H & \textbf{100.00} & \textbf{100.00} & \textbf{100.00} & 33.33\\
AAAC Problem I & \textbf{100.00} & \textbf{100.00} & \textbf{100.00} & 25.00\\
AAAC Problem J & \textbf{100.00} & \textbf{100.00} & \textbf{100.00} & 50.00\\
AAAC Problem K & \textbf{100.00} & \textbf{100.00} & 75.00 & 25.00\\
AAAC Problem L & \textbf{100.00} & \textbf{100.00} & \textbf{100.00} & 25.00\\
AAAC Problem M & 45.83 & 54.17 & \textbf{87.50} & 16.67\\
Average across ten problems & 86.89 & \textbf{88.49} & 87.60 & 25.64\\
& & & &\\
PAN12 Problem A & \textbf{100.00} & \textbf{100.00} & \textbf{100.00} & 33.33\\
PAN12 Problem C & 75.00 & 75.00 & \textbf{100.00} & 12.50\\
PAN12 Problem I & \textbf{92.86} & 85.71 & \textbf{92.86} & 35.71\\
Average across three problems & 89.29 & 86.90 & \textbf{97.62} & 27.18\\
\hline
\end{tabular}
\end{center}
\caption{\label{tab:competition_data}Classification performance of word network features on competition datasets. Along with our best test set accuracy values (second and third columns), we report here the best and lowest \emph{reported} accuracy values on the test set. Second column shows our results using pure word network features, and third column shows the results using word network features + raw term frequency of words. Best results for different problems (i.e., on different rows) are boldfaced.}
% The last column gives descriptions of our systems that achieved the accuracy values reported in the first column.
\end{table*}

With this view in mind, we went ahead and experimented with local features on nine different representative sets of words (cf. Section~\ref{subsec:wn_features}). We compared the performance of these features with the term frequency of representative words in the document (\emph{baseline feature set} extracted in Section~\ref{subsec:wn_features}). The results are shown in Table~\ref{tab:gutenberg_stop_multiclass} through Table~\ref{tab:gutenberg_mf_multiclass_mix}. Note that in each case, either vertex coreness or vertex neighborhood size (order 1) was the best-performing feature, and they significantly outperformed the baseline (term frequency)\footnote{We omitted the results on other local features for brevity.}. This shows the potential for using local word network features (as opposed to summary features) in authorship attribution problem.

Vertex clustering coefficient, however, almost always performed poorer than other local features and the term frequency baseline, and therefore seems rather unsuitable for authorship classification task. One reason behind this poor performance may lie in the fact that clustering coefficient smooths out fine-grain variations that are captured by frequentist measures like term frequency, vertex degree, vertex coreness, and vertex neighborhood size. And this smoothing of important variations may also lie at the heart of the poor performance exhibited by vertex clustering coefficient used in conjunction with nearest-neighbor methods like 1NN.

Note also that 1NN was the best classifier in most of the cases. This indicates a relationship between local features and 1NN. While using summary features, Logit Boost performed the best (Tables~\ref{tab:gutenberg_summary_multiclass} and~\ref{tab:gutenberg_summary_onevsall}). On using local features, 1NN came out to be the best performer. Moreover, in Table~\ref{tab:gutenberg_stop_multiclass}, Na\"{i}ve Bayes went well with vertex clustering coefficients (although the overall performance was low).

Furthermore, if we go from Table~\ref{tab:gutenberg_stop_multiclass} to Table~\ref{tab:gutenberg_mf_multiclass_mix}, we observe a general trend. As the number of words in our representative sets increases, so does the classification performance. A bigger list of stopwords performs substantially better than a smaller one, and the set of mood words always performs poorly, no matter what local feature we choose. This leads to another important finding of our study - stopwords (and most frequent words) are extremely useful and robust for authorship attribution of documents on a large scale. Similar findings have been reported in earlier research~\cite{Diederich:2003:AAS:776973.776982,menon-choi:2011:RANLP}. Moreover, raw term frequency as an additional feature is barely helpful if we consider the best results in each column. It improved performance in only two cases (Table~\ref{tab:gutenberg_stop_multiclass_mix} third column and Table~\ref{tab:gutenberg_mf_multiclass_mix} second column). One of them was marginal improvement.

A careful examination of Tables~\ref{tab:gutenberg_stop_multiclass} and~\ref{tab:gutenberg_mf_multiclass} reveals another interesting pattern. Note that vertex coreness was unable to beat the strong term frequency baseline when the representative set of words was too small (54 and 174 stopwords in Table~\ref{tab:gutenberg_stop_multiclass}, and $k$ = 100 and 200 in Table~\ref{tab:gutenberg_mf_multiclass}). This indicates a degree of sensitivity of the coreness measure to the number of words used to represent a document. Analyzing the relationship between these local features and the number of words in the representative set will be an interesting area for further research.

Combining local features with term frequency yields mixed results. Note from Tables~\ref{tab:gutenberg_stop_multiclass_mix} and~\ref{tab:gutenberg_mf_multiclass_mix} that clustering coefficient improves in almost all cases, coreness improves when the number of words in the representative set is small, and neighborhood size slightly goes down in terms of performance. Vertex degree shows mixed outcome - decreasing performance in general, with mood words being an exception.

Since we have many different types of features and several different representative sets of words, it is instructive to look into the relative performance of these features on one particular representative word set. Table~\ref{tab:ig_ranking_gutenberg_500_most_frequent_words} shows top 20 features ranked according to information gain on the whole Gutenberg dataset (3,036 documents and 142 classes). Here we considered most frequent 500 words as our representative set, and we took the term frequency, clustering coefficient, vertex degree, coreness, and neighborhood size of these words to create our feature vector. Each document is therefore represented as a feature vector of length 2,500. Note that among the top-ranked features, there are many local word network features alongside term frequency, which shows the relative competence of those features with respect to term frequency. Moreover, stopwords like \emph{until}, \emph{by}, \emph{several} and \emph{thus} appear several times among the top-ranked features, thereby broadly showing the power of stopwords in predicting authorial style.

\subsection{On Competition Data}
\label{subsec:aaac_pan12}

We obtained poor results using graph summary features on competition datasets, just like the Gutenberg data. On AAAC data, summary features gave best \emph{overall} test set accuracy of 26.25\% (using Logit Boost). Best overall test set accuracy reached 33.53\% on PAN12 data (using Logit Boost). While these results are much better than a na\"{i}ve random baseline (1.94\% $\pm$ 1.59\% on AAAC; 4.17\% $\pm$ 3.85\% on PAN12), we have obtained substantial gains by using local word network features on representative sets of words. Table~\ref{tab:competition_data} outlines the results on AAAC and PAN12 problems. Note that in eight out of the ten AAAC problems, and in two out of the three PAN12 problems, word network features performed on par with or better than the \emph{best reported result} (Table~\ref{tab:competition_data} columns one and three). Performance remained nearly the same when we added raw term frequency (Table~\ref{tab:competition_data} column two), showing once again that raw term frequency as an additional feature does not help much. Note from Table~\ref{tab:competition_data} that word network features sometimes performed \emph{better} than the best reported result (AAAC problems B, G, K), thereby showing the potential of these features in authorship attribution task. We achieved these results using simple graph features and term frequency, as opposed to more complex features like author profiles and cross-entropy~\cite{Juola:2006:AA:1373450.1373451}.

\section{Conclusions and Future Work}
\label{sec:conclusion}

In this paper, we described a set of novel features derived from a word network representation of text documents. We used these features for the traditional NLP task of authorship attribution and found significant performance gains when these features are used on top of a reasonably large list of stopwords. We further showed the effectiveness of local graph features in large-scale and competition-type authorship attribution problems, along with pointing out the relatively poor performance of graph summary features and non-frequentist local features like vertex clustering coefficient. Moreover, by using an array of experiments, we were able to select a set of best-performing classifiers and a set of features on which they performed the best.

Future work consists of combining different types of local features to see if that improves performance, analyzing the relationship between size of representative word sets and the performance of local features, and using the one-vs-all classification scheme instead of the multiclass classification scheme to see if it changes performance.

\bibliographystyle{naaclhlt2013}
\bibliography{forpaper}

\end{document}